\documentclass{ecai}
\usepackage{graphicx}
\usepackage{latexsym}
\usepackage{amsmath}
\usepackage{subcaption}

\newcommand{\xnot}{{\sf xNot360}}

\usepackage{longtable}
\usepackage{hyperref}

\begin{document}

\begin{frontmatter}

\title{A negation detection assessment of GPTs:  analysis with the \xnot~dataset}

\author[A]{\fnms{Ha Thanh}~\snm{Nguyen}\orcid{0000-0003-2794-7010}\thanks{Corresponding Author. Email: nguyenhathanh@nii.ac.jp}}
\author[B]{\fnms{Randy}~\snm{Goebel}}
\author[C]{\fnms{Francesca}~\snm{Toni}}
\author[D]{\fnms{Kostas}~\snm{Stathis}}
\author[A]{\fnms{Ken}~\snm{Satoh}}


\address[A]{National Institute of Informatics, Japan}
\address[B]{University of Alberta, Canada}
\address[C]{Imperial College London, UK}
\address[D]{Royal Holloway, University of London, UK }

\begin{abstract}
Negation is a fundamental aspect of natural language, playing a critical role in communication and comprehension.
Our study assesses the negation detection performance of Generative Pre-trained Transformer (GPT) models, specifically GPT-2, GPT-3, GPT-3.5, and GPT-4. We focus on the identification of negation in natural language using a zero-shot prediction approach applied to our custom \xnot{} dataset. Our approach examines sentence pairs labeled to indicate whether the second sentence negates the first. Our findings expose a considerable performance disparity among the GPT models, with GPT-4 surpassing its counterparts and GPT-3.5 displaying a marked performance reduction. The overall proficiency of the GPT models in negation detection remains relatively modest, indicating that this task pushes the boundaries of their natural language understanding capabilities. We not only highlight the constraints of GPT models in handling negation but also emphasize the importance of logical reliability in high-stakes domains such as healthcare, science, and law. 
\end{abstract}

\end{frontmatter}

\section{Introduction}
\label{sec:intro}

Negation is a fundamental aspect of natural language, playing a critical role in communication and comprehension. Despite the impressive performance of state-of-the-art pre-trained language models on numerous tasks, they often struggle to handle negation correctly \cite{hosseini-etal-2021-understanding, hossain-etal-2022-analysis, hossain-etal-2020-analysis, geiger-etal-2020-neural}. This deficiency highlights the need for a deeper understanding of negation in natural language processing (NLP) and motivates the development of improved models that can more accurately incorporate negation in their representations.

Existing studies suggest that popular natural language understanding corpora contain few negations compared to general-purpose English, and the negations present are often unimportant \cite{hossain-etal-2022-analysis}. In fact, it is often possible to ignore negations and still make correct predictions. But state-of-the-art transformer models trained on such corpora perform significantly worse on instances containing negation, especially when the negations are crucial for comprehension \cite{hossain-etal-2020-analysis, geiger-etal-2020-neural}.

To address this challenge, researchers have proposed augmenting language models with an unlikelihood objective based on negated generic sentences from raw text corpora \cite{hosseini-etal-2021-understanding}. This approach has led to a reduction in mean top 1 error rate on negated datasets, as well as some improvements on general negated natural language inference benchmarks. Meanwhile, the creation of new benchmarks and datasets has focused on negation, such as Monotonicity NLI (MoNLI) \cite{geiger-etal-2020-neural}, which aims to push the boundaries of natural language understanding models when it comes to dealing with negation.

\begin{figure}[ht]
\centering
\includegraphics[width=.5\textwidth]{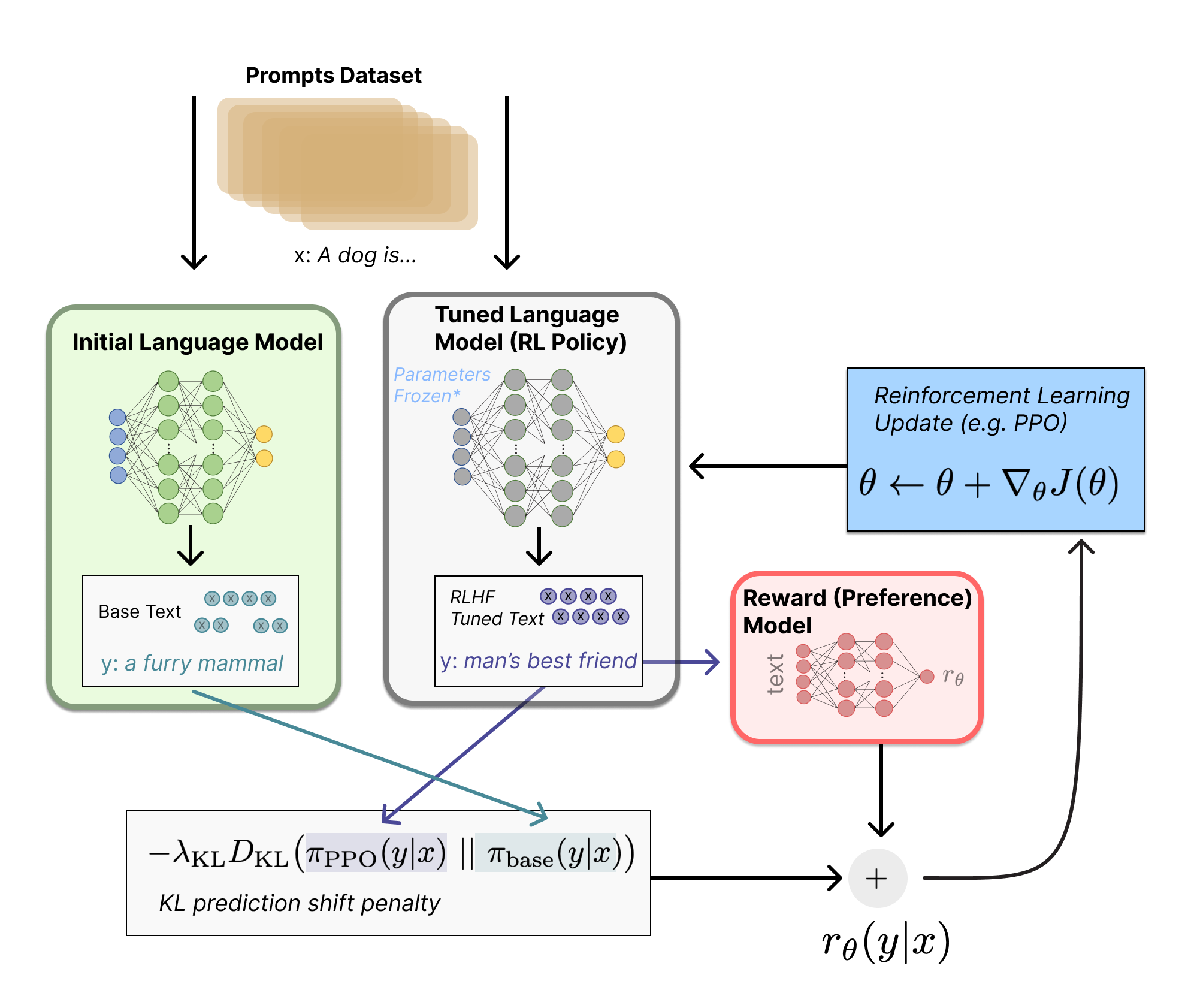}
\caption{The RLHF mechanism increases model alignment with user intent rather than relying on ground truth. Source image: \href{https://huggingface.co/blog/rlhf}{https://huggingface.co/blog/rlhf}}
\label{fig:rlhf}
\end{figure}

However, the lack of standard annotation schemes and guidelines for capturing negation in existing corpora poses a significant challenge \cite{jimenez-zafra-etal-2020-corpora}. To make progress in negation treatment, it is crucial to develop standardized approaches for annotating negation information in different languages and to improve the compatibility of corpora. By doing so, we can facilitate the development of more robust language models capable of accurately handling negation and, ultimately, enhance their performance across various NLP tasks.


Throughout the development of the Generative Pre-trained Transformer (GPT) models, the automated understanding of language has evolved. GPT-2 \cite{radford2019language} demonstrated impressive zero-shot learning capabilities on language modeling tasks, while GPT-3 \cite{brown2020language} significantly increased the trained model size,  and achieved strong performance on numerous NLP tasks, including translation and question-answering. The GPT-3.5 model (a.k.a. ChatGPT) \cite{ouyang2022training} introduced reinforcement learning from human feedback (RLHF), which is a ``fine-tuning'' process aimed at better aligning the model with user intent. Lastly, GPT-4 \cite{openai2023gpt4} is a multimodal model capable of handling image and text inputs, and exhibits human-level performance on various professional and academic benchmarks.

Utilizing RLHF, as illustrated in Figure \ref{fig:rlhf}, generative models adapt their outputs to achieve the highest reward in terms of user satisfaction. Nevertheless, constantly optimizing for user satisfaction will not always yield favorable outcomes. The adage 'Bitter pills may have blessed effects' is an appropriate metaphor, suggesting that offering users more palatable information can occasionally undermine the potency of the message, a phenomenon labelled in the GPT domain as an \textit{hallucination}~\cite{ouyang2022training}.
While RLHF is an interesting and intelligent approach to improving model alignment with user intent, it will face challenges if the majority of 
users are not genuinely concerned with the logical consistency of accumulating language use. This semantic challenge is in addition to the as yet unassessed challenge of scaling this technique to continually improve such models.

Natural language is not based on accuracy but rather on effectiveness, as evidenced by the inherent ambiguity present in human communication. This ambiguity is not a flaw but a feature of natural language.
Natural language, as opposed to formal languages like mathematics or programming languages, is primarily designed to facilitate communication and convey meaning between humans. As such, it prioritizes effectiveness and adaptability in diverse contexts, rather than strict accuracy and precision. This is evident in the way natural language often relies on context, ambiguity, and shared knowledge between the speaker and the listener to convey information efficiently and effectively.  It is also confirmed that interaction with natural language is a foundational component of explanation, in order to improve communication and resolve ambiguity \cite{make3040045}.

In many cases, natural language provides multiple ways to express the same idea, and speakers can choose the phrasing that best suits the situation or the intended audience. This flexibility is essential for efficient communication, as it enables speakers to adapt their language to different social, cultural, and situational contexts. However, this flexibility can also introduce ambiguity and imprecision.  
Within this context, it is crucial to provide benchmark GPT models on complex logic tasks such as negation detection. Doing so can contribute valuable insights into the limitations and potential advances of future research within this realm. This kind of focus provides important highlights on logical reliability in high-stakes domains such as law and healthcare (e.g. see \cite{Rabelo_2022}, \cite{10.1145/2651444}, \cite{mullenbach-etal-2018-explainable}, \cite{lertvittayakumjorn2021supporting}, \cite{nguyen2022attentive}), 
where an accurate and consistent understanding of negation is essential for effective communication and decision-making.

To evaluate the performance of these models, one contribution of this work is the creation of the \textit{eXploring Negation Over Text with 360 samples} (\xnot{}) dataset\footnote{https://huggingface.co/datasets/nguyenthanhasia/xNot360}. This dataset is specifically designed for assessing negation detection capabilities of language models across various contexts and domains. The \xnot{} dataset contains a diverse set of sentences with negations and their corresponding non-negated counterparts, which supports a comprehensive evaluation of the models' performance on this task. The motivation behind creating the \xnot{} dataset is to provide a more challenging benchmark for negation detection, taking into account the subtleties and complexities of negation in natural language.

There are also two other key contributions of our work. First, we benchmark and compare the performance of GPT-2, GPT-3, GPT-3.5, and GPT-4 on the \xnot{} dataset using a zero-shot prediction approach;  and we provide insights into the strengths and weaknesses of these models with respect to negation detection, which inform future research on improvements in natural language understanding. 
Additionally, numerous language models have proven their effectiveness, such as LLaMA~\cite{touvron2023llama}, Bard\footnote{https://bard.google.com/}, Jurassic-2\footnote{https://www.ai21.com/blog/introducing-j2}, Claude\footnote{https://www.anthropic.com/index/introducing-claude}. However, GPT models have detailed documentation, a development history, and have been tested on various benchmarks. As a result, we have chosen to focus on the GPT model family as the subject of our research in this paper.

The rest of our paper is organized as follows: Section 2 discusses related work on negation detection and benchmarking GPT models in similar tasks. Section 3 provides details of our methodology, including construction of the \xnot{} dataset, the zero-shot prediction approach, and evaluation metrics. Section 4 presents the experimental results and performance comparisons of the GPT models. Section 5 discusses the implications of our experiments and potential limitations of the study. Finally, Section 6 concludes the paper and summarizes our intended contributions to the field of natural language understanding.

\section{Previous Studies on GPT Models}

Large-scale pre-trained language models (LLMs) have significantly advanced the field of NLP. These models have demonstrated remarkable performance on a wide range of tasks, including text generation, question answering, and translation. 
Among these, Generative Pre-trained Transformers (GPT) utilize a transformer architecture and have achieved state-of-the-art results across various NLP tasks. In recent years, researchers have been exploring the capabilities and limitations of these models in different application domains, including legal and medical reasoning.

One of the notable achievements of GPT models is their performance in legal tasks. For instance, GPT-4 has been shown to pass the Uniform Bar Examination (UBE) with a score significantly higher than the passing threshold for all jurisdictions \cite{katz2023gpt}. This demonstrates the potential of GPT models in providing support for legal services in society. However, we argue that a model with only average logical capabilities can still achieve high scores on the Uniform Bar Examination (UBE), while also acknowledging the potential limitations of such a model in real-world legal practice. 

\begin{figure}[ht]
\centering
\includegraphics[width=.4\textwidth]{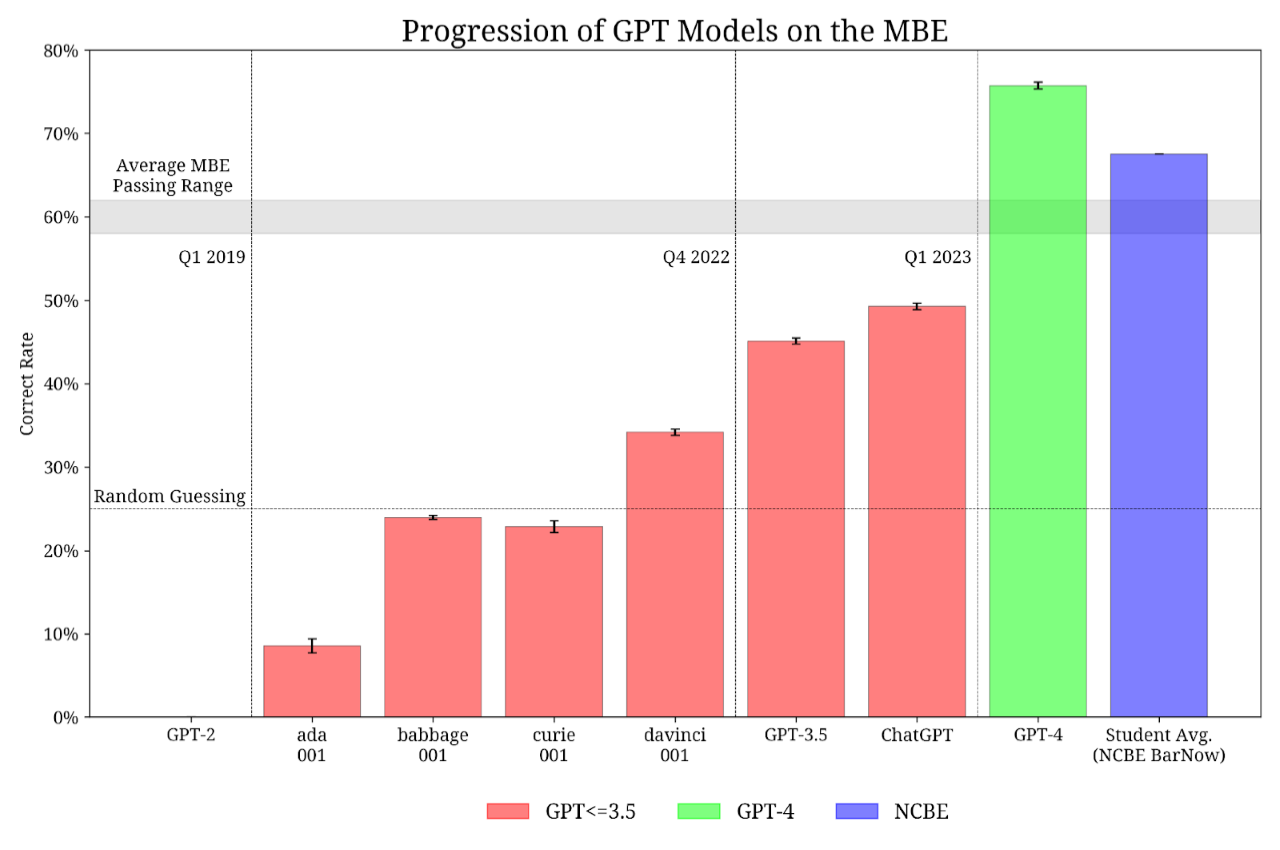}
\caption{GPT-4 far exceeds its predecessor models and even exceeds the average performance of law students on the MBE test. \cite{katz2023gpt}}
\label{fig:gpt4_mbe}
\end{figure}

First, it is essential to consider the structure of the UBE, which is based on the Multistate Bar Examination (MBE), the Multistate Essay Examination (MEE), and the Multistate Performance Test (MPT). The MBE focuses on multiple-choice questions, which primarily measure a candidate's ability to recall and apply legal principles. A model with average logical capabilities may excel in this section by exploiting its extensive knowledge base to identify correct answers.
Second, the MEE assesses an individual's ability to communicate legal information effectively through written essays. A model with average logical capabilities could perform well in this section, especially by demonstrating strong writing skills and a solid understanding of legal concepts. However, the model's potential limitations in logical reasoning might not be fully exposed in the MEE, as essay questions may not always delve deeply into complex logical analysis.
Lastly, the MPT evaluates practical lawyering skills through performance-based tasks, such as drafting legal documents or analyzing case files. While a model with average logical capabilities might struggle with some aspects of these tasks, it could still achieve high scores by showcasing its ability to synthesize and present relevant information.

As a result, a model with average logical capabilities may indeed achieve high scores on the UBE, {\it but this does not guarantee its effectiveness in real-world legal practice}. The GPT model's limitations in logical reasoning become more apparent when dealing with complex legal issues and real-life situations, which typically require more than just knowledge recall and basic reasoning. Therefore, while GPT models may demonstrate impressive performance on the UBE, their practical application in the legal profession should be approached with caution.

\begin{figure}
\centering
\includegraphics[width=.5\textwidth]{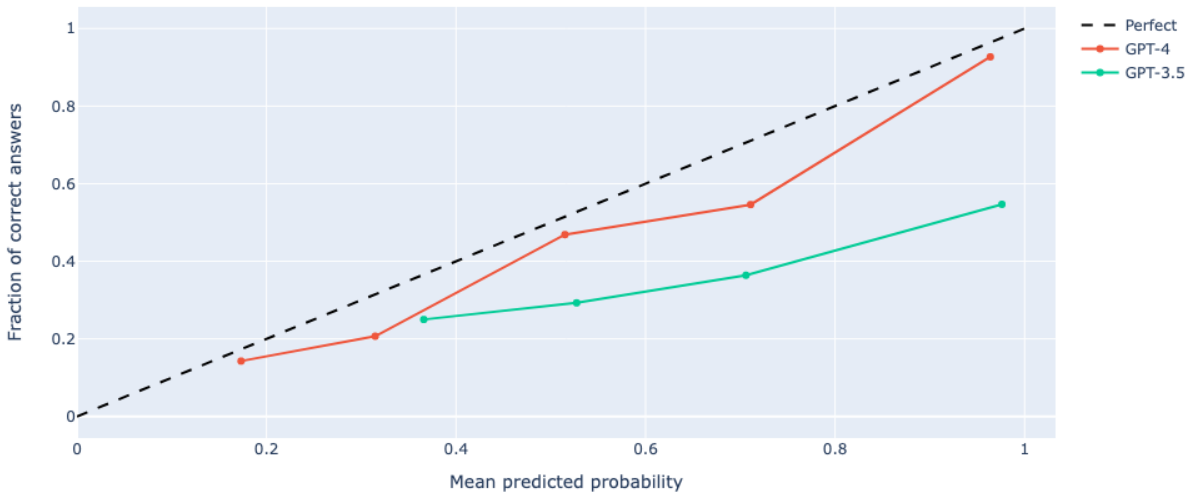}
\caption{GPT-4 is close to perfect calibration in the USMLE test. \cite{nori2023capabilities}}
\label{fig:gpt4_usmle}
\end{figure}

Nori et al. \cite{nori2023capabilities} present a comprehensive evaluation of GPT-4, a state-of-the-art large language model (LLM), on medical competency examinations and benchmark datasets, including the United States Medical Licensing Examination (USMLE), as shown in Figure \ref{fig:gpt4_usmle}. While GPT-4 demonstrates impressive performance on these tasks, it is important to note that these results do not confirm that model's logical capabilities.
Instead, GPT-4's high scores may be attributed to its pattern-matching and memorization abilities, as the task is not designed to test the model's logical reasoning skills specifically.

Several arguments and pieces of evidence can be put forth to support this claim. Firstly, the tasks evaluated in medical competency examinations mainly focus on the model's ability to recall and apply medical knowledge, rather than testing its logical reasoning or understanding of complex relationships in medical scenarios. This means that GPT-4's success on these tasks may {\it not} directly translate to its ability to perform logical reasoning in the medical domain.
Secondly, the paper highlights the risks of erroneous generations and hallucinations, where the model generates incorrect or misleading information. This implies that, despite its impressive performance, GPT-4 will still struggle with understanding the underlying logic and reasoning behind certain medical concepts and scenarios,
particularly in tasks such as differential diagnosis, treatment recommendations, and interpreting complex medical images. 
The potential consequences of such errors in high-stakes medical applications underscore the need for a better understanding of the model's logical capabilities.
Lastly, the evaluation metrics used in Nori's paper, such as accuracy and calibration, do not necessarily capture the model's logical proficiency. These metrics primarily focus on the model's ability to generate correct answers, but they do not reflect its understanding of the logical relationships and constraints involved in medical decision-making.

Given these concerns, a more effective approach to improve the model's logical capabilities would be to first develop robust methods for measuring and evaluating its logical reasoning performance. This could involve designing tasks and benchmarks specifically geared towards assessing the model's understanding of logical relationships, as well as creating evaluation metrics that better capture its proficiency in logical reasoning. As a result, researchers can gain valuable insights into the model's strengths and weaknesses, paving the way for targeted improvements and more reliable applications of LLMs in high-stakes domains.

Liu et al. \cite{liu2023evaluating} conducted a comprehensive evaluation of GPT-4's performance on various logical reasoning tasks using multiple datasets. They employed popular benchmarks such as LogiQA and ReClor, as well as newly-released datasets like AR-LSAT, to test the model's capabilities in multi-choice reading comprehension and natural language inference tasks that require logical reasoning. Additionally, they constructed an out-of-distribution dataset for logical reasoning to assess the robustness of ChatGPT and GPT-4.

The experimental results showed that ChatGPT significantly outperformed the RoBERTa LLM fine-tuning method on most logical reasoning benchmarks. With early access to the GPT-4 API, the authors were able to conduct extensive experiments on the GPT-4 model, revealing that it achieved even higher performance on most logical reasoning datasets. Both ChatGPT and GPT-4 performed relatively well on well-known datasets like LogiQA and ReClor. However, their performance dropped considerably when dealing with newly-released and out-of-distribution datasets, indicating that logical reasoning remains challenging for these models, particularly on out-of-distribution and natural language inference datasets.

Upon closer inspection of the errors made by GPT-4 on the LogiQA dataset, the authors categorized four of them as logical errors, three as scope errors, and the remaining three as the inability to resolve semantic ambiguity. Although this limited sample of errors does not conclusively demonstrate GPT-4's incompetence in handling reasoning questions, it does suggest that further examination and analysis are needed.

Given the findings from Liu et al.'s study, it becomes evident that there is a need for a dedicated dataset focusing specifically on negation detection. While the existing datasets and benchmarks provide a broad overview of the models' logical reasoning capabilities, a specialized dataset for negation detection would enable researchers to delve deeper into this particular aspect of natural language understanding. In response to this need, we introduce our custom \xnot{} dataset, designed explicitly for evaluating GPT models' proficiency in negation detection. Our goal is to to contribute valuable insights to improving the precision of the models and eliminating ambiguity in the negation context.

\section{Dataset and Experimental Settings}

\subsection{\xnot{} Dataset}
The \textit{eXploring Negation Over Text with 360 samples}   (\xnot{}) dataset is motivated by our observation of the poor performance of large language models (LLMs) in generating and recognizing negation when evaluated on abductive reasoning tasks.

Abductive reasoning \cite{walton2014abductive,bhagavatula2020abductive,Hunter_2022,liang2022visual} plays a crucial role in fields such as law and medicine, as it enables professionals to work through complex and often ambiguous cases by considering multiple hypotheses and ruling out those that do not align with the observed evidence, ultimately driving them toward the most plausible explanation.
The ``360'' in the name signifies a comprehensive, all-encompassing approach to negation detection, akin to a 360-degree view of the problem. This emphasizes both the dataset's purpose and its importance in evaluating LLMs' capabilities in handling negation across various contexts and sentence structures.

The dataset is motivated by our observation of the poor performance of large language models (LLMs) in generating and recognizing negation when evaluated on abductive reasoning tasks. Although these models can easily use the word ``not'' or negation phrases to generate or recognize simple negated sentences, they often struggle with more complex sentence structures.
Creating logical negation is not only challenging for machines, but it can also be difficult for humans. This suggests that RLHF, a method based on human voting, may not be useful. So, also for this reason, we use classical logic as a guideline when constructing this dataset.

We have noticed that conditional statements are particularly prone to errors when negated. For example, the sentences ``If I study hard, I will pass the bar exam'' and ``If I do not study hard, I will not pass the bar exam'' may initially appear to negate each other, but they do not.
Let $A$ represent ``I study hard'' and $B$ represent ``I will pass the bar exam.'' The first sentence can be represented as $A \to B$, which is equivalent to $\lnot A \lor B$. The second sentence can be represented as $\lnot A \to \lnot B$, which is equivalent to $A \lor \lnot B$. To demonstrate that the first sentence is not the negation of the second sentence, we can construct a truth table as in Table~\ref{tab:truth_value}:

\begin{table}
\centering
\begin{tabular}{|c|c|c|c|c|}
\hline
A & B & $\lnot A \lor B$ & $A \lor \lnot B$ & $\lnot A \lor B \neq A \lor \lnot B$ \\
\hline
T & T & T & T & F \\
T & F & F & T & T \\
F & T & T & F & T \\
F & F & T & T & F \\
\hline
\end{tabular}
\caption{Truth table showing where the first sentence and the second sentence are not negations of each other, as their logical expressions differ. This highlights the complexity of negation in natural language and underscores the need for a dataset that can improve the evaluation of LLMs on this task. \label{tab:truth_value}}
\end{table}

To construct the \xnot{} dataset, we first designed various sentence templates and then negated some components to create positive and negative labels as discussed earlier. The dataset comprises 360 samples, including 180 positive samples, which are sentence pairs that negate each other, and 180 negative samples, which are sentence pairs that do not negate each other. The length of the sentences in each pair ranges from 5 to 20 words. 

\begin{figure*}
\centering
\includegraphics[width=0.8\textwidth]{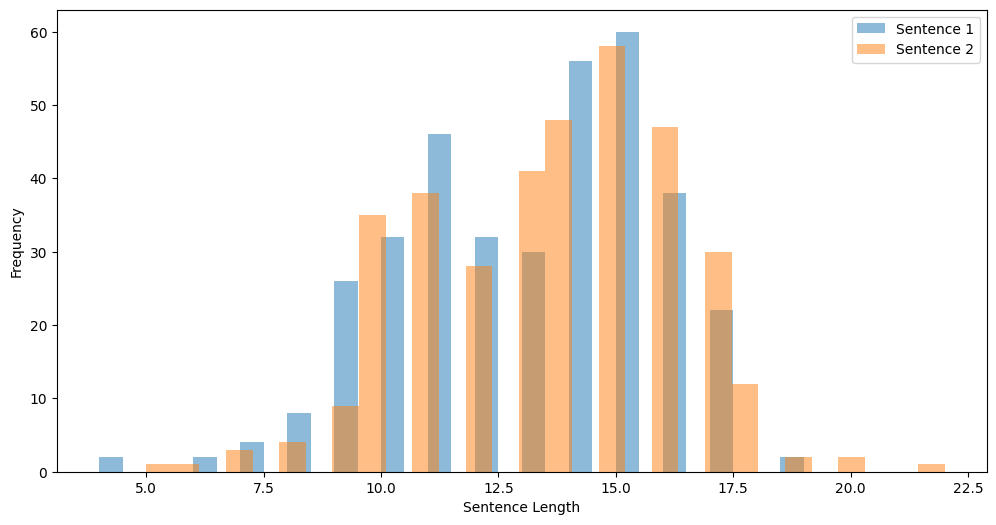}
\caption{Distribution of sentence lengths in the \xnot{} dataset.}
\label{fig:length_distribution}
\end{figure*}

As seen in Figure \ref{fig:length_distribution}, the distribution of sentence lengths approximates a Gaussian distribution, highlighting the variability in sentence complexity within the dataset. This ensures a diverse range of samples for evaluating the negation detection capabilities of LLMs.

\subsection{Experimental Settings}

We evaluate the GPT-2, GPT-3, GPT-3.5, and GPT-4 models using a zero-shot prediction approach. The models are given a prompt and required to predict whether the second sentence in a pair negates the first sentence or not. We use the following prompt for all models:

\begin{quote}
    Sentence 1: \textit{\{sentence1\}} \\
    Sentence 2: \textit{\{sentence2\}} \\
    Is Sentence 2 a negation of Sentence 1? (1: Yes, 0: No)
\end{quote}

For GPT-2, we utilize the \textit{zero-shot-classification} pipeline provided by HuggingFace\footnote{\href{https://huggingface.co/tasks/zero-shot-classification}{https://huggingface.co/tasks/zero-shot-classification}}, as GPT-2 is publicly available. The other models (GPT-3, GPT-3.5, and GPT-4) are accessed through the OpenAI API\footnote{\href{https://openai.com/blog/openai-api}{https://openai.com/blog/openai-api}} using the \textit{text-davinci-003}, \textit{gpt-3.5-turbo}, and \textit{gpt-4} endpoints, respectively.

To evaluate the models' performance, we use accuracy, precision, recall, and F1-score as the primary metrics. We also plot confusion matrices to visualize the distribution of true positives, true negatives, false positives, and false negatives for each model. To gain a deeper understanding of the models' limitations, we analyze the errors made by the models on specific sentence pairs, highlighting their weaknesses and failure patterns.

\section{Experimental Results}

Here we report the experimental results for each GPT model, including accuracy and F1-score. The performance of the models is compared and discussed, as well as the analysis of common misclassified sentences and potential reasons for the errors.

\subsection{Model Performance}

\begin{table}
\centering
\begin{tabular}{|c|c|c|c|c|}
\hline
\textbf{Model} & \textbf{Accuracy} & \textbf{F1-score} & \textbf{Precision} & \textbf{Recall} \\ \hline
GPT-2           & 0.5000            & 0.6667            & 0.5000             & 1.0000           \\ \hline
GPT-3           & 0.6056            & 0.6913            & 0.5679             & 0.8833           \\ \hline
GPT-3.5         & 0.4306            & 0.2705            & 0.3762             & 0.2111           \\ \hline
GPT-4           & 0.7833            & 0.7706            & 0.8187             & 0.7278           \\ \hline
\end{tabular}
\caption{Performance comparison of GPT models on the \xnot{} dataset.}
\label{tab:model_performance}
\end{table}

Table \ref{tab:model_performance} summarizes the performance of the GPT models on our \xnot{} dataset. GPT-4 outperforms the other models in both major metrics, achieving an accuracy of 0.7833, an F1-score of 0.7706, a precision of 0.8187, and a recall of 0.7278. On the other hand, GPT-3.5 shows a significant performance dip compared to its counterparts, with an accuracy of 0.4306, an F1-score of 0.2705, a precision of 0.3762, and a recall of 0.2111. GPT-3 exhibits moderate performance, with an accuracy of 0.6056, an F1-score of 0.6913, a precision of 0.5679, and a recall of 0.8833. Lastly, GPT-2 has the lowest accuracy (0.5000) but achieves the highest recall (1.0000).

\begin{figure*}
\centering
\includegraphics[width=.8\textwidth]{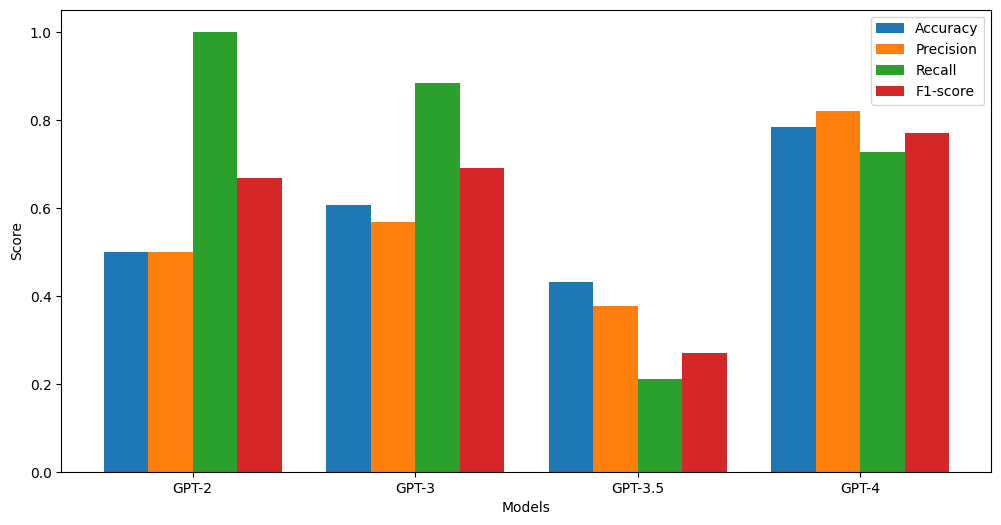}
\caption{Performance chart of GPT models on the \xnot{} dataset. The chart displays a sinusoidal-like pattern, highlighting the differences in performance among the models.}
\label{fig:performance_chart}
\end{figure*}

To provide a visual representation of the performance variations among the four GPT models, we have plotted a performance chart that illustrates the fluctuating nature of their accuracy, F1-score, precision, and recall metrics. The chart exhibits a sinusoidal-like pattern, indicating the disparities in performance across the models.

As shown in Figure \ref{fig:performance_chart}, the performance of the GPT models varies considerably. GPT-4 exhibits the highest performance across major metrics, while GPT-3.5 demonstrates a noticeable dip in performance. GPT-3, on the other hand, shows moderate performance and GPT-2 has the lowest accuracy but the highest recall.
This sinusoidal pattern in the performance chart underscores the inconsistencies in the GPT models' ability to handle negation, emphasizing the need for further research and refinement in this area.

\subsection{Confusion Matrix Analysis}
\begin{figure*}
\centering
\begin{subfigure}{.24\textwidth}
  \centering
  \includegraphics[width=\linewidth]{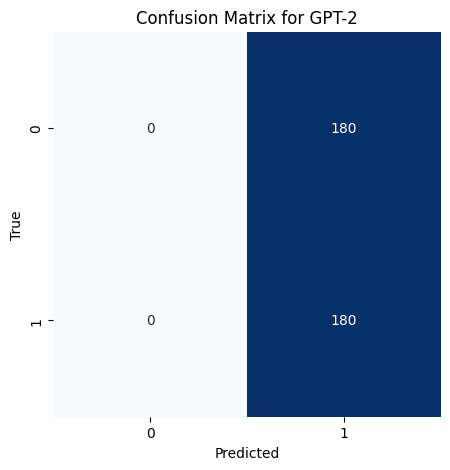}
  \caption{GPT-2}
  \label{fig:gpt2_matrix}
\end{subfigure}
\begin{subfigure}{.24\textwidth}
  \centering
  \includegraphics[width=\linewidth]{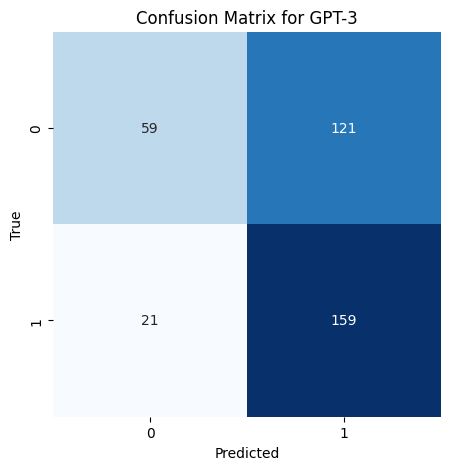}
  \caption{GPT-3}
  \label{fig:gpt3_matrix}
\end{subfigure}
\begin{subfigure}{.24\textwidth}
  \centering
  \includegraphics[width=\linewidth]{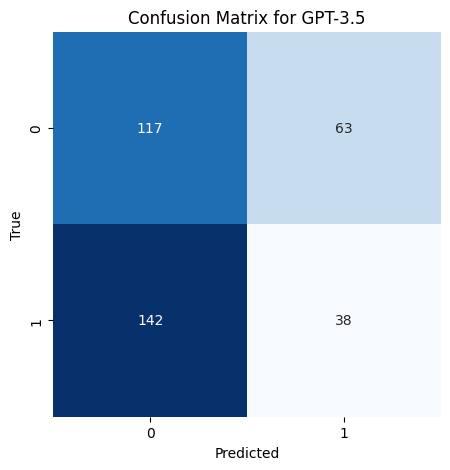}
  \caption{GPT-3.5}
  \label{fig:gpt3_5_matrix}
\end{subfigure}
\begin{subfigure}{.24\textwidth}
  \centering
  \includegraphics[width=\linewidth]{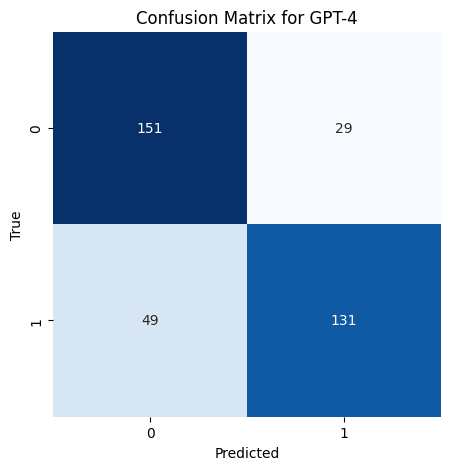}
  \caption{GPT-4}
  \label{fig:gpt4_matrix}
\end{subfigure}
\caption{Confusion matrices of GPT-2, GPT-3, GPT-3.5, and GPT-4 predictions on \xnot{}.}
\label{fig:confusion_matrices}
\end{figure*}

\subsubsection{GPT-2}




Figure \ref{fig:gpt2_matrix} shows that GPT-2 predicts all labels to be 1, resulting in an accuracy of exactly 50\%. According to the original paper, this model ``achieves state-of-the-art results on 7 out of 8 tested language modeling datasets in a zero-shot setting.'' However, it appears to have no innate understanding of negation. Our samples were generated in 2023, while the web data from 2019 does not contain these examples; hence, GPT-2 cannot synthesize the data and make accurate predictions. Interestingly, 50\% is not the worst performance. We will delve deeper into this phenomenon with GPT-3.5 and explore an intriguing hypothesis involving RLHF in the following sections.

Figure \ref{fig:gpt3_matrix} shows that GPT-3 still tends to predict label 1 more often, which may reflect a general similarity with its predecessor. There could be something in the training process of these models that makes them lean towards positive predictions rather than negative ones. Comparing the accuracy of 0.5 for GPT-2 to 0.6056 for GPT-3, it seems like a significant improvement. However, in terms of F1-score, the improvement is quite modest, with only a 0.02 difference compared to a random guess (0.67 for GPT-2 and 0.69 for GPT-3). This suggests that GPT-3 is not a powerful model for negation detection. To put it more intuitively, if \xnot{} were an exam, GPT-2 would score a grade D while GPT-3 would score a grade C.

Figure \ref{fig:gpt3_5_matrix} provides the confusion matrix of GPT-3.5 on \xnot{}, which shows that this model tends to predict more label 0's than label 1's, which is than its two predecessors. It is worth noting that GPT-3.5 is the first model to implement RLHF, a mechanism that allows users to vote on the model's responses. Additionally, it is a model with a series of ethical and legal constraints.

What both surprises and intrigues us is that this model's performance on \xnot{} is even lower than a random guess, specifically with 0.4306 accuracy and 0.2705 F1-score. This aligns with our initial assumption that for a ``small-sized'' model (compared to the problems it needs to understand), aligning the parameters to cater to user preferences or to comply with unnatural constraints may impact its performance. This is not a new issue in AI and is similar to the degradation caused by an ineffective heuristic rule or overfitting to a less general dataset. Returning to the \xnot{} exam analogy, ChatGPT scores an F grade.

Figure \ref{fig:gpt4_matrix}, the confusion matrix of GPT-4 on \xnot{}, displays a more balanced distribution of predictions for both label 0 and label 1 compared to its predecessors. This suggests that GPT-4 has a better understanding of negation and is more proficient in detecting it.

With an accuracy of 0.7833 and an F1-score of 0.7706, GPT-4 significantly outperforms the previous models, demonstrating a substantial improvement in negation detection. Although far from perfect, this progress indicates that the model is learning from its training data and better grasping the concept of negation. In the context of our \xnot{} exam analogy, GPT-4 scores a grade B, reflecting its enhanced ability to handle negation.

This improvement may be attributed to the advancements in model architecture, training data, and optimization techniques employed in GPT-4. However, it is essential to continue investigating and refining the model's performance for negation detection, especially in high-stakes domains such as healthcare, science, and law, where logical reliability and understanding of negation are critical.

\subsection{Error Patterns}

\begin{table*}[!ht]
\centering
\begin{tabular}{|l|l|l|}
\hline
\textbf{No.} & \textbf{Sentences} & \textbf{Labels} \\ \hline
1 & Sentence 1: Drinking water frequently helps maintain proper hydration & True: 0\\
  & Sentence 2: Not drinking water frequently helps maintain proper hydration &  Predicted: 1 \\ \hline
2 & Sentence 1: If a person is found guilty beyond a reasonable doubt, they can be convicted. & True: 0 \\
  & Sentence 2: If a person is not found guilty beyond a reasonable doubt, they cannot be convicted. &  Predicted: 1 \\ \hline
3 & Sentence 1: If you drive in the designated lane, you will avoid penalties. & True: 0 \\
  & Sentence 2: If you do not drive in the designated lane, you will avoid penalties. & Predicted: 1 \\ \hline
4 & Sentence 1: Studying hard will result in good grades & True: 0\\
  & Sentence 2: Not studying hard will result in good grades & Predicted: 1 \\ \hline
5 & Sentence 1: If you pay child support, the child's needs will be financially covered. & True: 0\\
  & Sentence 2: If you do not pay child support, the child's needs will be financially covered. & Predicted: 1 \\ \hline
6 & Sentence 1: If you mediate your divorce, the process will be less adversarial. & True: 0\\
  & Sentence 2: If you do not mediate your divorce, the process will be less adversarial. &  Predicted: 1 \\ \hline
7 & Sentence 1: If you establish paternity, you will have legal rights to the child. & True: 0\\
  & Sentence 2: If you do not establish paternity, you will have legal rights to the child. &  Predicted: 1 \\ \hline
8 & Sentence 1: If you maintain a safe distance from other vehicles, you will reduce the risk of collisions. & True: 0 \\
  & Sentence 2: If you do not maintain a safe distance from other vehicles, you will reduce the risk of collisions. & Predicted: 1 \\ \hline
9 & Sentence 1: Saving money consistently will help you build wealth & True: 0\\
  & Sentence 2: Not saving money consistently will help you build wealth &  Predicted: 1 \\ \hline
10 & Sentence 1: Practicing mindfulness reduces stress & True: 0\\
   & Sentence 2: Not practicing mindfulness reduces stress &  Predicted: 1 \\ \hline
11 & Sentence 1: If a person is found guilty of kidnapping, they can be sentenced to prison. & True: 0\\
   & Sentence 2: If a person is not found guilty of kidnapping, they cannot be sentenced to prison. &  Predicted: 1 \\ \hline
12 & Sentence 1: If a person is granted parole, they can be released from prison early. & True: 0\\
   & Sentence 2: If a person is not granted parole, they cannot be released from prison early. &  Predicted: 1 \\ \hline
\end{tabular}
\caption{Sentences that all models misclassified.}
\label{tab:misclassified_sentences}
\end{table*}

Table \ref{tab:misclassified_sentences} lists 12 sentence pairs for which all models failed. Since GPT-2 always predicts label 1, we tried removing the condition ``GPT-2 predicts incorrectly'' from the filter, resulting in 18 sentence pairs that GPT-3, GPT-3.5, and GPT-4 all predicted incorrectly. From these results, we observe that most of the erroneous predictions arise from the models reliance on negation cues in one part of a conditional statement to determine the negation status of the entire sentence.

In this list, there are sentence pairs that even humans find challenging to determine their negation relationship. For example, sentence pair 9: ``Saving money consistently will help you build wealth'' and ``Not saving money consistently will help you build wealth.'' At first glance, relying on the template ``Not+sentence'' as the negation of ``sentence'' might lead to an incorrect decision.

These examples further support the appropriateness of using logic as a guiding principle for creating a negation on text dataset like our \xnot{}. This is not to suggest that a logical treatment is always straightforward to apply, as the transformation of natural language to logic is a foundational problem.  But some studies in the application of negation rules to legal reasoning, we have already seen practical improvements (e.g., \cite{Kim_2015}).  So by employing logical rules, we can at least ensure that our dataset effectively tests a model's ability to understand and handle negation in text.

\section{Future Works}
\subsection{Evaluating other generative models}
\begin{figure}
\centering
\includegraphics[width=.43\textwidth]{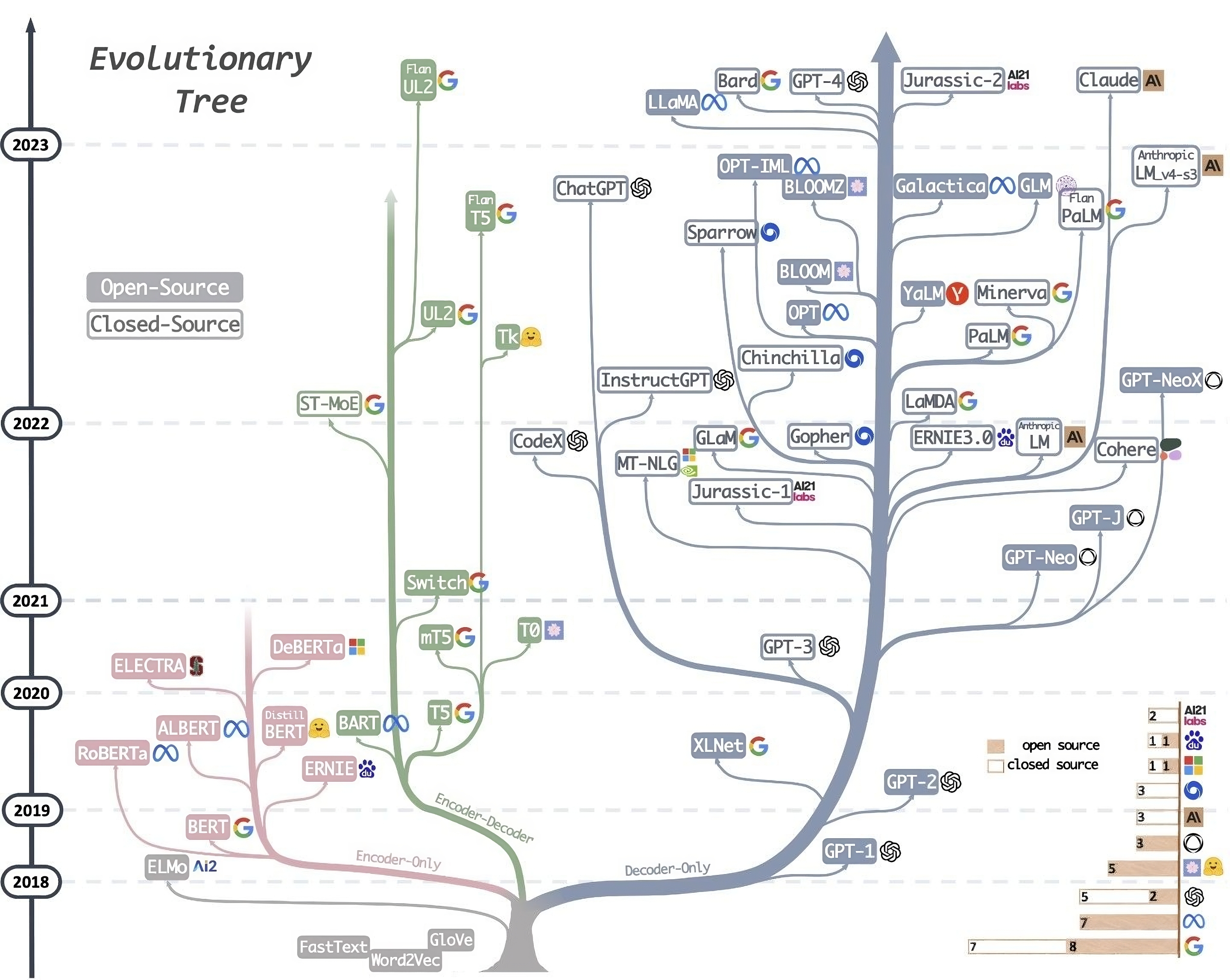}
\caption{The evolutionary tree of modern LLMs traces the development of language models in recent years and highlights some of the most well-known models \cite{yang2023harnessing}. }
\label{fig:evolutionary_tree}
\end{figure}

In addition to GPTs, we plan to evaluate other generative models using a similar approach in the near future. With the growing interest in generative models, we anticipate that models with approaches similar to GPT will become increasingly prevalent. Figure \ref{fig:evolutionary_tree} presents an evolutionary tree of modern large language models (LLMs), showcasing the development of language models in recent years and highlighting some of the most well-known models. By assessing these models in the context of negation detection, we aim to contribute to a more comprehensive understanding of their capabilities and limitations.

\subsection{Expanding the dataset with more negation types}
To gain a more comprehensive view of the capabilities of various models in handling negation, we plan to create an expanded dataset that includes a broader range of negation types. Based on the information provided in the paper \cite{hafissatou2021analysis}, we will incorporate different negation constructions, such as affixal negation (e.g., dis-, un-, anti-, -less), standard negation (e.g., not, n't), negative imperatives (e.g., do not, don't), negative quantifiers (e.g., not everyone, not many), and negative indefinite pronouns (e.g., nobody, no one). By integrating these diverse negation structures into our dataset, we aim to present a more challenging and comprehensive evaluation of the models' abilities to handle negation in NLP.

For instance, we may include sentences with affixal negation, such as ``The patient's condition was incurable,'' where the negation is embedded within the word ``incurable'' through the use of the negative prefix ``in-''. Similarly, we could incorporate sentences with negative quantifiers, such as ``Few judges ruled in favor of the defendant,'' where the negation is expressed within the subject of the sentence.

By expanding our dataset to encompass a wider variety of negation types, we hope to provide a more detailed and engaging assessment of the logical reasoning capabilities of different NLP models.

\section{Discussions and Conclusions}

We have presented a systematic study on the performance of GPT-2, GPT-3, GPT-3.5, and GPT-4 models in negation detection using a zero-shot prediction approach. Our analysis used our custom \xnot{} dataset, designed specifically to assess the models' proficiency in understanding and handling negation across various contexts and sentence structures.

Our findings revealed considerable performance disparities amongst the GPT models. While GPT-4 exhibited the highest proficiency in negation detection, GPT-3.5 displayed a marked dip in performance compared to its counterparts. The overall competence of negation detection remains relatively modest, suggesting that this task pushes the boundaries of their natural language understanding capabilities.

Through our investigation, we have not only highlighted the limitations of current state-of-the-art pre-trained language models in dealing with negation but also emphasized the significance of logical reliability in high-stakes domains such as healthcare, science, and law. Our work offers valuable insights that can inform future research directions within this realm.

As part of our ongoing efforts to improve natural language understanding and develop more robust language models capable of more accurately representing and handling negation, future work would focus on refining existing evaluation methodologies or creating new benchmarks tailored specifically for assessing logical reasoning capabilities. Additionally, we recommend exploring techniques to further enhance model alignment with user intent and investigating novel training methods that prioritize logical consistency without compromising other aspects of model performance.

By continuing to shed light on the challenges associated with negation detection in NLP and by contributing resources like our \xnot{} dataset to the research community, we hope to inspire further advancements in this critical aspect of natural language understanding.

\bibliography{ecai,anthology}

\begin{thebibliography}{10}

\bibitem{bhagavatula2020abductive}
Chandra Bhagavatula, Ronan~Le Bras, Chaitanya Malaviya, Keisuke Sakaguchi, Ari
  Holtzman, Hannah Rashkin, Doug Downey, Wen{-}tau Yih, and Yejin Choi,
  `Abductive commonsense reasoning', in {\em 8th International Conference on
  Learning Representations, {ICLR} 2020}, Addis Ababa, Ethiopia, (2020).
  OpenReview.net.

\bibitem{brown2020language}
Tom Brown, Benjamin Mann, Nick Ryder, Melanie Subbiah, Jared~D Kaplan, Prafulla
  Dhariwal, Arvind Neelakantan, Pranav Shyam, Girish Sastry, Amanda Askell,
  et~al., `Language models are few-shot learners', {\em Advances in neural
  information processing systems}, {\bf 33},  1877--1901, (2020).

\bibitem{geiger-etal-2020-neural}
Atticus Geiger, Kyle Richardson, and Christopher Potts, `Neural natural
  language inference models partially embed theories of lexical entailment and
  negation', in {\em Proceedings of the Third BlackboxNLP Workshop on Analyzing
  and Interpreting Neural Networks for NLP}, pp. 163--173, Online, (November
  2020). Association for Computational Linguistics.

\bibitem{hafissatou2021analysis}
KANE Hafissatou et~al., `An analysis of negation in english', {\em
  International Journal of English Language Studies}, {\bf 3}(4),  100--106,
  (2021).

\bibitem{hossain-etal-2022-analysis}
Md~Mosharaf Hossain, Dhivya Chinnappa, and Eduardo Blanco, `An analysis of
  negation in natural language understanding corpora', in {\em Proceedings of
  the 60th Annual Meeting of the Association for Computational Linguistics
  (Volume 2: Short Papers)}, pp. 716--723, Dublin, Ireland, (May 2022).
  Association for Computational Linguistics.

\bibitem{hossain-etal-2020-analysis}
Md~Mosharaf Hossain, Venelin Kovatchev, Pranoy Dutta, Tiffany Kao, Elizabeth
  Wei, and Eduardo Blanco, `An analysis of natural language inference
  benchmarks through the lens of negation', in {\em Proceedings of the 2020
  Conference on Empirical Methods in Natural Language Processing (EMNLP)}, pp.
  9106--9118, Online, (November 2020). Association for Computational
  Linguistics.

\bibitem{hosseini-etal-2021-understanding}
Arian Hosseini, Siva Reddy, Dzmitry Bahdanau, R~Devon Hjelm, Alessandro
  Sordoni, and Aaron Courville, `Understanding by understanding not: Modeling
  negation in language models', in {\em Proceedings of the 2021 Conference of
  the North American Chapter of the Association for Computational Linguistics:
  Human Language Technologies}, pp. 1301--1312, Online, (June 2021).
  Association for Computational Linguistics.

\bibitem{Hunter_2022}
Anthony Hunter, `Understanding enthymemes in deductive argumentation using
  semantic distance measures', {\em Proceedings of the {AAAI} Conference on
  Artificial Intelligence}, {\bf 36}(5),  5729--5736, (jun 2022).

\bibitem{jimenez-zafra-etal-2020-corpora}
Salud~Mar{\'\i}a Jim{\'e}nez-Zafra, Roser Morante, Mar{\'\i}a~Teresa
  Mart{\'\i}n-Valdivia, and L.~Alfonso Ure{\~n}a-L{\'o}pez, `Corpora annotated
  with negation: An overview', {\em Computational Linguistics}, {\bf 46}(1),
  1--52, (2020).

\bibitem{katz2023gpt}
Daniel~Martin Katz, Michael~James Bommarito, Shang Gao, and Pablo Arredondo,
  `Gpt-4 passes the bar exam', {\em Available at SSRN 4389233}, (2023).

\bibitem{make3040045}
Mi-Young Kim, Shahin Atakishiyev, Housam Khalifa~Bashier Babiker, Nawshad
  Farruque, Randy Goebel, Osmar~R. ZaÃ¯ane, Mohammad-Hossein Motallebi,
  Juliano Rabelo, Talat Syed, Hengshuai Yao, and Peter Chun, `A multi-component
  framework for the analysis and design of explainable artificial
  intelligence', {\em Machine Learning and Knowledge Extraction}, {\bf 3}(4),
  900--921, (2021).

\bibitem{Kim_2015}
Mi-Young Kim, Ying Xu, and Randy Goebel, `Legal question answering using
  ranking svm and syntacitc/semantic similarity', {\em New Frontiers in
  Artificial Intelligence, Lecture Notes in Computer Science}, {\bf 9067},
  244--258, (2015).

\bibitem{10.1145/2651444}
Mi-Young Kim, Ying Xu, Osmar~R. Zaiane, and Randy Goebel, `Recognition of
  patient-related named entities in noisy tele-health texts', {\em ACM Trans.
  Intell. Syst. Technol.}, {\bf 6}(4), (jul 2015).

\bibitem{lertvittayakumjorn2021supporting}
Piyawat Lertvittayakumjorn, Ivan Petej, Yang Gao, Yamuna Krishnamurthy, Anna
  Van Der~Gaag, Robert Jago, and Kostas Stathis, `Supporting complaints
  investigation for nursing and midwifery regulatory agencies', in {\em
  Proceedings of the 59th Annual Meeting of the Association for Computational
  Linguistics and the 11th International Joint Conference on Natural Language
  Processing: System Demonstrations}, pp. 81--91, (2021).

\bibitem{liang2022visual}
Chen Liang, Wenguan Wang, Tianfei Zhou, and Yi~Yang, `Visual abductive
  reasoning', in {\em Proceedings of the IEEE/CVF Conference on Computer Vision
  and Pattern Recognition}, pp. 15565--15575, (2022).

\bibitem{liu2023evaluating}
Hanmeng Liu, Ruoxi Ning, Zhiyang Teng, Jian Liu, Qiji Zhou, and Yue Zhang,
  `Evaluating the logical reasoning ability of chatgpt and gpt-4', {\em arXiv
  preprint arXiv:2304.03439}, (2023).

\bibitem{mullenbach-etal-2018-explainable}
James Mullenbach, Sarah Wiegreffe, Jon Duke, Jimeng Sun, and Jacob Eisenstein,
  `Explainable prediction of medical codes from clinical text', in {\em
  Proceedings of the 2018 Conference of the North {A}merican Chapter of the
  Association for Computational Linguistics: Human Language Technologies,
  Volume 1 (Long Papers)}, pp. 1101--1111, New Orleans, Louisiana, (June 2018).
  Association for Computational Linguistics.

\bibitem{nguyen2022attentive}
Ha-Thanh Nguyen, Manh-Kien Phi, Xuan-Bach Ngo, Vu~Tran, Le-Minh Nguyen, and
  Minh-Phuong Tu, `Attentive deep neural networks for legal document
  retrieval', {\em Artificial Intelligence and Law},  1--30, (2022).

\bibitem{nori2023capabilities}
Harsha Nori, Nicholas King, Scott~Mayer McKinney, Dean Carignan, and Eric
  Horvitz, `Capabilities of gpt-4 on medical challenge problems', {\em arXiv
  preprint arXiv:2303.13375}, (2023).

\bibitem{openai2023gpt4}
OpenAI.
\newblock Gpt-4 technical report, 2023.

\bibitem{ouyang2022training}
Long Ouyang, Jeffrey Wu, Xu~Jiang, Diogo Almeida, Carroll Wainwright, Pamela
  Mishkin, Chong Zhang, Sandhini Agarwal, Katarina Slama, Alex Ray, et~al.,
  `Training language models to follow instructions with human feedback', {\em
  Advances in Neural Information Processing Systems}, {\bf 35},  27730--27744,
  (2022).

\bibitem{Rabelo_2022}
Juliano Rabelo, Randy Goebel, Mi-Young Kim, Yoshinobu Kano, Masaharu Yoshioka,
  and Ken Satoh, `Overview and discussion of the competition on legal
  information extraction/entailment ({COLIEE}) 2021', {\em The Review of
  Socionetwork Strategies}, {\bf 16}(1),  111--133, (feb 2022).

\bibitem{radford2019language}
Alec Radford, Jeffrey Wu, Rewon Child, David Luan, Dario Amodei, Ilya
  Sutskever, et~al., `Language models are unsupervised multitask learners',
  {\em OpenAI blog}, {\bf 1}(8), ~9, (2019).

\bibitem{touvron2023llama}
Hugo Touvron, Thibaut Lavril, Gautier Izacard, Xavier Martinet, Marie-Anne
  Lachaux, Timoth{\'e}e Lacroix, Baptiste Rozi{\`e}re, Naman Goyal, Eric
  Hambro, Faisal Azhar, et~al., `Llama: Open and efficient foundation language
  models', {\em arXiv preprint arXiv:2302.13971}, (2023).

\bibitem{walton2014abductive}
Douglas Walton, {\em Abductive reasoning}, University of Alabama Press, 2014.

\bibitem{yang2023harnessing}
Jingfeng Yang, Hongye Jin, Ruixiang Tang, Xiaotian Han, Qizhang Feng, Haoming
  Jiang, Bing Yin, and Xia Hu, `Harnessing the power of llms in practice: A
  survey on chatgpt and beyond', {\em arXiv preprint arXiv:2304.13712}, (2023).

\end{thebibliography}
\end{document}